%% file: acl_latex.tex
\pdfoutput=1

\documentclass[11pt]{article}

\usepackage[final]{acl}

\usepackage{times}
\usepackage{latexsym}

\usepackage[T1]{fontenc}

\usepackage[utf8]{inputenc}

\usepackage{microtype}

\usepackage{inconsolata}

\usepackage{graphicx}

\usepackage{amsmath}
\usepackage{amssymb}
\usepackage{multirow}
\usepackage{multicol}
\usepackage{graphicx}
\usepackage{enumitem}
\usepackage{hyperref}
\usepackage{tikz}
\usepackage{booktabs}
\usepackage{CJKutf8}

\usepackage{arydshln}

\newcommand*{\circled}[1]{\lower.7ex\hbox{\tikz\draw (0pt, 0pt)%
    circle (.5em) node {\makebox[1em][c]{\scriptsize #1}};}}

\title{Reference-Free Post-Training of Open Large Language Models for Multilingual Machine Translation}

\author{
  \textbf{Chris Han},
  \textbf{Pengzhi Gao}\thanks{~Corresponding author: \texttt{gaopengzhi@xiaomi.com}.},
  \textbf{Pei Fu},
  \textbf{Jian Luan}
\\
  Xiaomi Inc., Beijing, China
}

\begin{document}

\maketitle

\begin{abstract}

We study reference-free post-training for multilingual machine translation with open large language models. Starting from the supervised-finetuned MiLMMT-46-v0.1 models~\cite{shang2026scalingmodeldatamultilingual}, we apply Group Relative Policy Optimization (GRPO)~\cite{shao2024deepseekmathpushinglimitsmathematical} with a reward that averages two reference-free quality estimation models and is gated by language identification. We then linearly interpolate the supervised fine-tuning (SFT) and reinforcement learning (RL) model checkpoints to obtain MiLMMT-46-v1.0. Across 46 languages, the resulting models consistently improve translation quality over their SFT counterparts, outperform strong recent open baselines including Seed-X~\cite{cheng2025seedxbuildingstrongmultilingual}, HY-MT2~\cite{zheng2026hymt2technicalreport}, and TranslateGemma~\cite{finkelstein2026translategemmatechnicalreport}, and achieve leading reference-free scores against evaluated proprietary systems such as Google Translate, Gemini~3 Pro, and GPT-5. We further investigate on-policy distillation (OPD) and find that it reaches, but does not surpass, the quality frontier achieved by RL with checkpoint interpolation. We release the models and code to facilitate future research.\footnote{Models are released at \url{https://huggingface.co/collections/xiaomi-research/milmmt-46}. Codes are released at \url{https://github.com/xiaomi-research/gemmax}.}

\end{abstract}

\section{Introduction}
Open large language models (LLMs), including the Gemma and Qwen families~\cite{gemmateam2026gemma4technicalreport,yang2025qwen3technicalreport}, have recently become strong foundations for multilingual machine translation (MT). Through continual pretraining and supervised finetuning (SFT), recent work has adapted these models into capable multilingual translation systems. In particular, the MiLMMT-46-v0.1 models~\cite{shang2026scalingmodeldatamultilingual} achieve strong performance across $46$ languages, providing a strong open starting point for further post-training.

Despite these advances, further improving multilingual MT with supervised learning remains constrained by the scarcity and uneven coverage of high-quality parallel data, particularly for low-resource languages and non-English-centric translation directions. By contrast, source-side text is far more abundant and easier to collect across languages. This gap motivates reference-free post-training methods that can exploit source-side data without aligned target sentences.

Reinforcement learning (RL) provides a natural framework for this setting because it can optimize sequence-level rewards on model-generated translations rather than token-level likelihood against references. Group Relative Policy Optimization (GRPO)~\cite{shao2024deepseekmathpushinglimitsmathematical} has recently been applied to multilingual translation with carefully designed rewards, as exemplified by Tower+~\cite{rei2025towerbridginggeneralitytranslation} and HY-MT2~\cite{zheng2026hymt2technicalreport}. Reference-free quality estimation (QE) models can score a source sentence and its candidate translation without a reference, but imperfect estimators may encourage reward hacking~\cite{liu-etal-2026-mending}. A language-identification gate can suppress wrong-language outputs, while SFT--RL checkpoint interpolation can mitigate reward-induced drift and preserve behavior learned during SFT. These considerations motivate our combination of reference-free RL with checkpoint interpolation.

Building on the MiLMMT-46-v0.1 models, we study this approach across three model scales and $46$ languages. Our experiments show that reference-free RL consistently improves learned quality metrics over SFT and compares favorably with strong recent open and proprietary systems, while checkpoint interpolation provides a controllable trade-off between reference-free quality and reference-based performance. We further investigate on-policy distillation (OPD)~\cite{agarwal2024policy,lu2025onpolicydistillation} as an alternative post-training approach and find that it reaches, but does not surpass, the quality frontier achieved by RL with checkpoint interpolation. We release the resulting MiLMMT-46-v1.0 models and code to facilitate future research.

\section{Methodology}
\label{sec:method}

Starting from the supervised-finetuned MiLMMT-46-v0.1 models, we use Group Relative Policy Optimization (GRPO) to optimize a reference-free reward based on two quality estimation models and a language-identification gate. We then linearly interpolate the SFT and RL checkpoints to retain the behavior learned during supervised finetuning while incorporating the improvements from RL.

\paragraph{Training objective.} We adopt GRPO for the RL stage. For each source sentence $x$, we sample a group of $G$ candidate translations $\{y_1, \dots, y_G\}$ from the old policy $\pi_{\theta_{\text{old}}}$. Let $R_i$ denote the reward assigned to candidate $y_i$, and let $\pi_{\text{ref}}$ denote the fixed SFT reference policy. We maximize the following objective:
\begin{equation*}
\small
\begin{aligned}
\mathcal{J}(\theta) = \; & \mathbb{E}_{x,\, \{y_i\}} \bigg[ \tfrac{1}{G}\textstyle\sum_{i=1}^{G} \tfrac{1}{|y_i|} \sum_{t=1}^{|y_i|} \Big\{ \min\big[ r_{i,t}\,\hat{A}_{i,t}, \\[-2pt]
& \operatorname{clip}(r_{i,t}, 1{-}\epsilon, 1{+}\epsilon)\,\hat{A}_{i,t} \big] - \beta\, \mathbb{D}_{\text{KL}}(\pi_\theta \| \pi_{\text{ref}}) \Big\} \bigg],
\end{aligned}
\end{equation*}
where
\[
r_{i,t}(\theta)
=
\frac{
\pi_\theta(y_{i,t} \mid x,y_{i,<t})
}{
\pi_{\theta_{\mathrm{old}}}(y_{i,t} \mid x,y_{i,<t})
}
\]
is the token-level importance ratio, $\hat{A}_{i,t}$ is the group-normalized advantage computed from the rewards $\{R_i\}_{i=1}^{G}$ of the sampled translations, $\epsilon$ controls the clipping range, and $\beta$ controls the Kullback--Leibler (KL) penalty that regularizes the RL policy toward the SFT reference policy.

\paragraph{Reward design.} Given a source sentence $x$, a candidate translation $y$, and the intended target language $\ell$, we predict the language of $y$, denoted by $\hat{\ell}(y)$, using OpenLID-v3~\cite{fedorova-etal-2026-openlid}. Let $s_{\mathrm{X}}$ and $s_{\mathrm{K}}$ denote the scores assigned to $(x,y)$ by XCOMET\footnote{\url{https://huggingface.co/Unbabel/XCOMET-XXL}}~\cite{guerreiro-etal-2024-xcomet} and COMETKiwi\footnote{\url{https://huggingface.co/Unbabel/wmt23-cometkiwi-da-xxl}}~\cite{rei-etal-2023-scaling}, respectively. We define the reward as
\begin{equation*}
R(x,y,\ell) =
\begin{cases}
\tfrac{1}{2}\big(s_{\mathrm{X}} + s_{\mathrm{K}}\big),
& \hat{\ell}(y)=\ell, \\
0, & \text{otherwise}.
\end{cases}
\end{equation*}
This language-identification gate prevents fluent but wrong-language outputs from receiving high QE rewards, mitigating a known form of reward hacking~\cite{liu-etal-2026-mending}.

\paragraph{RL data.} We derive the RL dataset from the MiLMMT SFT data by retaining each example's source sentence and translation direction while discarding its reference translation. This yields $263982$ instances spanning $192$ translation directions. Because GRPO relies on within-group reward variation, rollout groups with nearly identical rewards provide little training signal. We therefore sample $G$ translations for each instance using the RL rollout configuration, score them with our reward function, and retain instances whose group mean $\mu_x$ and standard deviation $\sigma_x$ satisfy $0.30 < \mu_x < 0.95$ and $\sigma_x \ge 0.05$. The mean bounds remove instances with very low or high average rewards, while the standard-deviation threshold removes groups with little reward variation. This leaves $31572$ instances, randomly split into $30572$ training and $1000$ validation instances. A detailed breakdown by translation direction is provided in Appendix~\ref{app:rl_data_distribution}.

\paragraph{SFT--RL checkpoint interpolation.} Let $\theta_{\mathrm{SFT}}$ denote the parameters of MiLMMT-46-v0.1 and $\theta_{\mathrm{RL}}$ the parameters obtained after RL. To reduce reward-induced drift while retaining the gains from RL, we linearly interpolate the two checkpoints:
\begin{equation*}
\theta_{\alpha}
=
\alpha\theta_{\mathrm{SFT}}
+
(1-\alpha)\theta_{\mathrm{RL}},
\qquad \alpha \in [0,1].
\end{equation*}
Here, $\alpha$ controls the contribution of the SFT checkpoint, while $1-\alpha$ controls that of the RL checkpoint. The interpolation requires no additional training. We refer to $\theta_{\mathrm{RL}}$ as MiLMMT-46-v0.1-RL and the selected interpolated checkpoint as MiLMMT-46-v1.0.

\section{Experimental Setup}

\paragraph{Datasets.} We conduct experiments on $46$ languages spanning a broad linguistic spectrum, with detailed language information summarized in Table~\ref{tab:langs}. We evaluate multilingual translation performance on the FLORES+ \cite{nllb-24} and WMT24++ \cite{deutsch-etal-2025-wmt24} benchmarks. For WMT24++, we adopt the English source sentences and exclude those marked as low quality for reference-free evaluation.

\paragraph{Baselines.}\label{sec:baselines} We compare against strong proprietary systems, including Google Translate, Gemini 2.5/3 Pro, and GPT-5, as well as the large-scale multilingual NMT model NLLB \cite{nllbteam2022languageleftbehindscaling}. In addition, we compare against several strong open-source multilingual translation models:
\begin{itemize}[leftmargin=*]
\item Tower-Plus \cite{rei2025towerbridginggeneralitytranslation}: Gemma2/Qwen2.5-based models for multilingual translation and general-purpose tasks across $27$ languages.
\item GemmaX2-28 \cite{cui-etal-2025-multilingual}: Gemma2-based models designed for multilingual machine translation across $28$ languages.
\item Seed-X-Instruct/PPO \cite{cheng2025seedxbuildingstrongmultilingual}: Mistral-based models trained with instruction finetuning and reinforcement learning for multilingual machine translation across $28$ languages.
\item Hunyuan-MT, HY-MT1.5, and HY-MT2 \cite{zheng2025hunyuanmttechnicalreport,zheng2025hymt15technicalreport,zheng2026hymt2technicalreport}: Hunyuan-based multilingual translation models across $33$ languages.
\item TranslateGemma \cite{finkelstein2026translategemmatechnicalreport}: Gemma3-based models for high-quality translation across $55$ languages.
\end{itemize}

\paragraph{Training and checkpoint interpolation configurations.} We apply the same RL recipe at all three scales ($1$B, $4$B, and $12$B), using the \texttt{verl}\footnote{\url{https://github.com/verl-project/verl}} framework~\cite{sheng2025hybridflow} with vLLM-based~\cite{kwon2023efficient} rollouts. For each RL instance, we sample $G{=}8$ candidates and regularize the policy toward the SFT reference $\pi_{\text{ref}}$ with a low-variance KL penalty. After RL training, we interpolate the SFT and RL parameters to construct the final checkpoint. For the MiLMMT-46-v1.0 models, we set the interpolation coefficient to $\alpha=0.5$ at all three scales. We explore other values of $\alpha$ in the interpolation analysis. A complete list of hyper-parameters is provided in Appendix~\ref{app:hyperparams}.

\paragraph{Evaluation.} We use XCOMET and COMETKiwi as reference-free evaluators on both the FLORES+ and WMT24++ benchmarks. Both are $10$B-parameter models with high correlation with human judgments~\cite{freitag-etal-2023-results}. On FLORES+, we additionally report reference-based spBLEU and XCOMET scores. For each baseline, we evaluate on the subset of languages shared with our $46$-language model. Translations are generated using greedy decoding.

\input{main_results_table.tex}

\section{Experimental Results}

\subsection{Main Results}

Table~\ref{tab:main_results} reports reference-free results on WMT24++ and both reference-free and reference-based results on FLORES+. Since the compared systems differ in language coverage, each block uses the language subset shared by its included systems.

\paragraph{Effects of reference-free post-training.} Within the full $46$-language block, MiLMMT-46-v1.0 consistently improves over its supervised-finetuned MiLMMT-46-v0.1 counterpart across all three model scales. Averaged over the three model scales, XCOMET and COMETKiwi scores on WMT24++ improve by $2.75$ and $2.44$ points, respectively. On FLORES+, averaged over the four direction groups and three model scales, reference-based XCOMET, reference-free XCOMET, and COMETKiwi improve by $1.17$, $1.41$, and $1.17$ points, while spBLEU decreases by $1.21$ points. These results show that reference-free post-training consistently improves learned quality metrics, with a modest reduction in lexical overlap. Since spBLEU measures lexical overlap with a single reference translation and can penalize valid alternative wording, we view its modest decrease as less informative than the consistent gains in learned quality metrics.

\paragraph{Full $46$-language comparison.} MiLMMT-46-12B-v1.0 achieves the highest reference-free XCOMET and COMETKiwi scores on WMT24++ and across all four FLORES+ direction groups, outperforming the evaluated proprietary systems, including Google Translate, Gemini~2.5/3 Pro, and GPT-5. On reference-based XCOMET, it ranks among the top two systems in three of the four FLORES+ direction groups. Among the smaller models, MiLMMT-46-1B-v1.0 outperforms TranslateGemma-4B on every reported WMT24++ and FLORES+ metric despite using only one quarter as many parameters. Although its spBLEU remains below Google Translate and Gemini~2.5/3 Pro, MiLMMT-46-12B-v1.0 surpasses GPT-5 in three of the four FLORES+ direction groups and achieves a higher average across the four groups ($35.14$ vs.\ $34.94$). It also achieves a higher average spBLEU across the four groups than all evaluated external open-source baselines, including NLLB-54.5B and TranslateGemma-27B.

\paragraph{Comparisons on shared language subsets.} Across the $21$-, $26$-, $28$-, and $31$-language subsets, MiLMMT-46-12B-v1.0 ranks first in $15$ of the $16$ reference-based XCOMET comparisons across the four FLORES+ direction groups. It also achieves the best reference-free scores in nearly all comparisons against Tower-Plus, Seed-X, GemmaX2, and the HY-MT series. These results are not limited to the largest model. MiLMMT-46-4B-v1.0 remains competitive with substantially larger models, including Tower-Plus-72B and HY-MT2-30B-A3B.

\subsection{Analysis}
\label{sec:analysis}

\begin{figure}[t]
    \centering
    \includegraphics[width=\columnwidth,trim=55pt 285pt 55pt 285pt,clip]{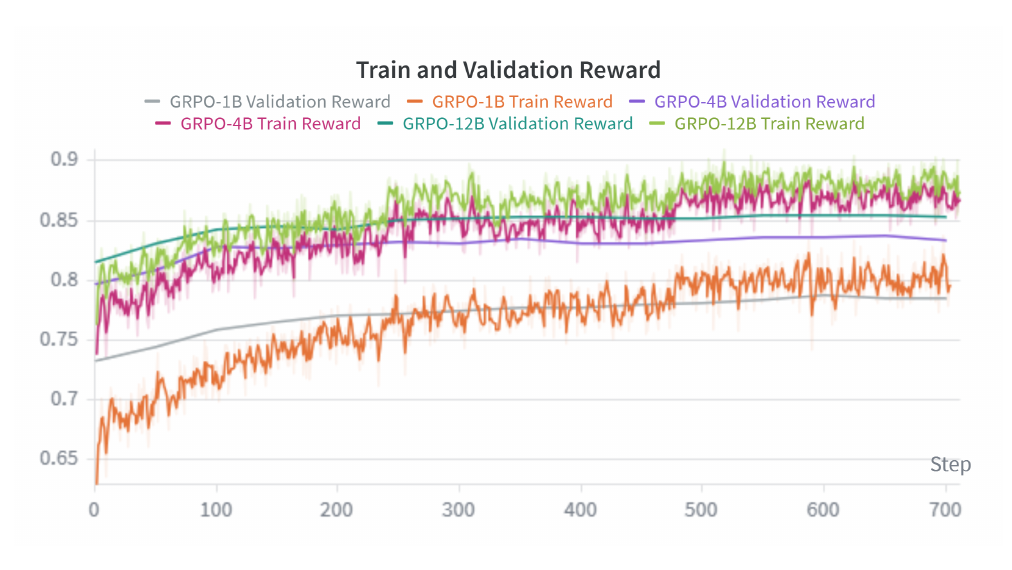}
    \caption{Training and validation rewards during GRPO post-training at the $1$B, $4$B, and $12$B scales. Validation rewards are evaluated every $50$ training steps.}
    \label{fig:reward_curve}
\end{figure}

\begin{figure*}[t]
    \centering
    \includegraphics[width=\textwidth]{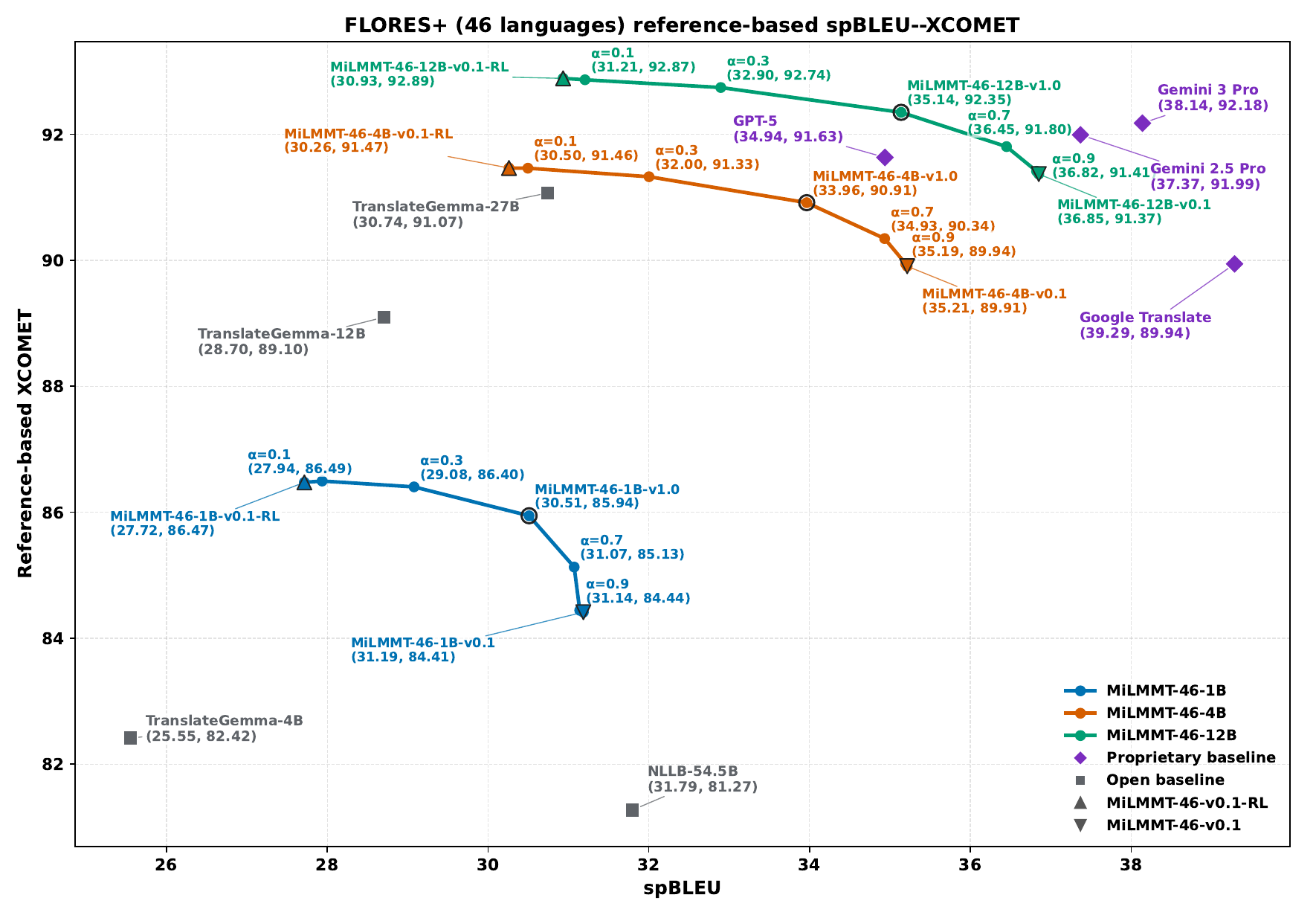}
    \caption{Trade-off between spBLEU and reference-based XCOMET under SFT--RL checkpoint interpolation on FLORES+, averaged over four translation-direction groups: \texttt{en} $\rightarrow$ \texttt{xx}, \texttt{xx} $\rightarrow$ \texttt{en}, \texttt{zh} $\rightarrow$ \texttt{xx}, and \texttt{xx} $\rightarrow$ \texttt{zh}. Each colored curve corresponds to one model scale, and $\alpha$ denotes the SFT weight. The MiLMMT-46-v0.1-RL and MiLMMT-46-v0.1 checkpoints form the two endpoints of each curve. Circled points indicate the selected MiLMMT-46-v1.0 checkpoints at $\alpha=0.5$. External systems are included for comparison.}
    \label{fig:interpolation_tradeoff}
\end{figure*}

\paragraph{Reference-free post-training gains persist across model scales.}
Figure~\ref{fig:reward_curve} shows that training and validation rewards increase early and then stabilize at all three model scales. Comparing MiLMMT-46-v1.0 with its SFT counterpart v0.1 on WMT24++, reference-free XCOMET improves by $2.87$, $2.59$, and $2.79$ points at the $1$B, $4$B, and $12$B scales, respectively. COMETKiwi shows the same trend. These results indicate that reference-free post-training remains effective across model scales, including on top of the strongest $12$B SFT model, rather than being limited to smaller models.

\paragraph{Interpolation as a controllable quality trade-off.}
Figure~\ref{fig:interpolation_tradeoff} shows a consistent trade-off between spBLEU and reference-based XCOMET across all three model scales. As the SFT weight $\alpha$ increases, the interpolated checkpoints move toward the SFT endpoint: spBLEU increases monotonically, while reference-based XCOMET decreases. The selected $\alpha=0.5$ checkpoints provide a favorable balance along these curves. Relative to the RL endpoints, they recover $2.79$, $3.70$, and $4.21$ spBLEU points at the $1$B, $4$B, and $12$B scales, respectively, while reducing reference-based XCOMET by only $0.53$, $0.56$, and $0.54$ points. Increasing $\alpha$ beyond $0.5$ yields diminishing returns in spBLEU. For the $12$B model, increasing $\alpha$ from $0.5$ to $0.7$ gains another $1.31$ spBLEU points at a cost of $0.55$ XCOMET, whereas increasing it from $0.7$ to $0.9$ gains only $0.37$ spBLEU points at a further cost of $0.39$ XCOMET. We therefore fix $\alpha=0.5$ across model scales as a simple operating point that recovers a substantial portion of the spBLEU lost during RL while retaining most of the improvement in reference-based XCOMET.

\section{On-Policy Distillation Analysis}
\label{sec:opd}

\begin{table}[t]
\centering\small
\setlength{\tabcolsep}{4pt}
\begin{tabular}{l | c | c c}
\hline
Model & WMT24++ & \multicolumn{2}{c}{FLORES+} \\
\hline
\hline
\textbf{1B} & & & \\
\quad MiLMMT-46-1B-v1.0           & 79.01 & 30.51 / 85.94 \\
\quad OPD (init. v0.1)             & 77.67 & 30.47 / 85.22 \\
\quad RL + OPD (init. v0.1)        & 78.25 & 30.49 / 85.67 \\
\quad OPD (init. v1.0)             & 77.86 & 30.39 / 85.42 \\
\hline
\textbf{4B} & & & \\
\quad MiLMMT-46-4B-v1.0           & 85.86 & 33.96 / 90.91 \\
\quad OPD (init. v0.1)             & 85.42 & 34.17 / 90.82 \\
\quad RL + OPD (init. v0.1)        & 85.85 & 33.98 / 90.98 \\
\quad OPD (init. v1.0)             & 85.37 & 34.16 / 90.79 \\
\hline
\end{tabular}
\caption{On-policy distillation from MiLMMT-46-12B-v1.0 into $1$B and $4$B students, averaged over $46$ languages. The v1.0 baseline is obtained through RL and checkpoint interpolation, and parenthetical labels indicate the student initialization. WMT24++ reports reference-free XCOMET, while FLORES+ reports spBLEU and reference-based XCOMET. RL + OPD adopts $\lambda=1$. Full per-group results and results for all tested values of $\lambda$ are provided in Appendix~\ref{app:opd_full}.}
\label{tab:opd}
\end{table}

We study whether on-policy distillation (OPD) can transfer the gains of a post-trained teacher to smaller students without running reinforcement learning separately at each model scale. We use MiLMMT-46-12B-v1.0 as the teacher and train $1$B and $4$B students using the same reference-free RL data described in Section~\ref{sec:method}. The students generate their own rollouts, and the teacher provides a distillation signal at the student-induced states.

Given a prompt $x$, a student rollout $y \sim \pi_\theta(\cdot \mid x)$, and a state $s_t = (x, y_{<t})$, OPD trains the student by minimizing the expected token-level divergence between the student policy and a teacher policy:
\[
\resizebox{\columnwidth}{!}{$\displaystyle
\mathcal{L}_{\mathrm{OPD}}(\theta)
=
\mathbb{E}_{x,\, y \sim \pi_\theta}
\!\left[
\tfrac{1}{|y|}
\textstyle\sum_{t=1}^{|y|}
D\!\big(
\pi_\theta(\cdot \mid s_t),
\pi_\phi(\cdot \mid s_t),
y_t
\big)
\right]
$}
\]
where $\pi_\phi$ denotes the teacher policy and $D$ is a per-token divergence signal. This objective admits two implementations. The distributional form, GKD~\cite{agarwal2024policy}, defines $D$ as a distribution-level KL and backpropagates through it directly. The policy-gradient form, PG-OPD~\cite{lu2025onpolicydistillation}, estimates the divergence from student-sampled tokens and optimizes the resulting signal as a reward through an RL update.

We adopt PG-OPD as our primary implementation to study OPD and RL under a unified objective. Unless stated otherwise, OPD refers to the policy-gradient form. We report GKD only as a control experiment in Appendix~\ref{app:opd_full}. Specifically, PG-OPD instantiates the per-token divergence for a student-sampled token $y_t$ as
\[
D_t
=
\operatorname{sg}\!\big(
\log \pi_\theta(y_t \mid s_t)
-
\log \pi_\phi(y_t \mid s_t)
\big),
\]
which is a single-sample estimator of the reverse KL divergence from the student policy to the teacher policy. Since $\mathcal{L}_{\mathrm{OPD}}$ averages $D_t$ over student-sampled tokens, minimizing it is equivalent to maximizing its negative. We therefore use $r_t=-D_t$ as the per-token reward and optimize it with the same GRPO objective described in Section~\ref{sec:method}. Here, $\operatorname{sg}(\cdot)$ denotes the stop-gradient operator, which prevents gradients from flowing through the reward computation while preserving its dependence on the teacher policy. We denote the resulting policy-gradient distillation loss by $\mathcal{L}_{\mathrm{distill}}$ and the GRPO loss from Section~\ref{sec:method} by $\mathcal{L}_{\mathrm{policy}}$.

We compare three training variants that differ in their objectives and student initialization. OPD (init. v0.1) optimizes $\mathcal{L}_{\mathrm{distill}}$ alone starting from MiLMMT-46-v0.1. RL+OPD (init. v0.1) combines the GRPO and distillation losses:
\begin{equation*}
\mathcal{L}
=
\mathcal{L}_{\mathrm{policy}}
+
\lambda\,\mathcal{L}_{\mathrm{distill}}
\end{equation*}
where $\lambda$ controls the distillation weight. OPD (init. v1.0) instead optimizes $\mathcal{L}_{\mathrm{distill}}$ alone starting from the corresponding MiLMMT-46-v1.0 student.

\paragraph{Transferring post-training gains through OPD.}
Table~\ref{tab:opd} compares the distilled students with the corresponding MiLMMT-46-v1.0 baselines. OPD (init. v0.1) closely matches the $4$B baseline on FLORES+, achieving $34.17$ versus $33.96$ spBLEU and $90.82$ versus $90.91$ reference-based XCOMET, while the $1$B student remains slightly behind in reference-based XCOMET ($85.22$ versus $85.94$). Overall, OPD from the post-trained $12$B teacher approaches the performance of the corresponding MiLMMT-46-v1.0 models, but does not consistently match or surpass them.

\paragraph{Distillation weight controls the quality trade-off.}
Varying $\lambda$ in RL + OPD produces a trade-off similar to that obtained through SFT--RL checkpoint interpolation in Section~\ref{sec:analysis}. At $4$B, reducing $\lambda$ from $1$ to $0.001$ increases reference-based XCOMET from $90.98$ to $91.47$ while decreasing spBLEU from $33.98$ to $30.20$, with intermediate values following the same trend. With $\lambda=1$, RL + OPD remains close to the corresponding MiLMMT-46-v1.0 baselines across both benchmarks and model scales. Initializing OPD from MiLMMT-46-v1.0 rather than MiLMMT-46-v0.1 likewise has little effect.

\paragraph{Overall finding.}
Across objectives, distillation weights, and student initializations, OPD and RL + OPD remain close to the quality frontier established by RL and checkpoint interpolation but do not extend it. We therefore view OPD as a robustness finding and an alternative way to transfer post-training gains to smaller models, rather than an improvement over RL and interpolation.

\section{Conclusion}
In this paper, we study reference-free post-training for multilingual machine translation with open large language models, analyzing GRPO, SFT--RL checkpoint interpolation, and on-policy distillation across the $1$B, $4$B, and $12$B model scales. Experiments covering $46$ languages on WMT24++ and FLORES+ show that language-gated quality estimation rewards consistently improve reference-free XCOMET and COMETKiwi, as well as reference-based XCOMET, although the resulting RL checkpoints exhibit lower spBLEU. Building on MiLMMT-46-v0.1, we develop MiLMMT-46-v1.0 by interpolating the SFT and RL checkpoints, recovering a substantial portion of the spBLEU lost during RL while retaining most of the gains in neural quality metrics. The resulting models compare favorably with strong recent open-source baselines, and MiLMMT-46-12B-v1.0 achieves leading reference-free results among the evaluated open and proprietary systems. Our on-policy distillation experiments further show that the gains of the post-trained $12$B teacher can be transferred to smaller students, although distillation does not improve on the trade-off achieved by RL and checkpoint interpolation. We hope our findings and released models support scalable and reference-free post-training for open multilingual translation systems. In future work, we will explore more robust and human-aligned reward models, extend reference-free post-training to multimodal multilingual translation, and study its behavior at larger model scales.


\bibliography{custom}

\appendix
\section{Distribution of RL Training Data}\label{app:rl_data_distribution}
After effective-data selection, the retained pool comprises
$31572$ source sentences ($30572$ for training and $1000$ held out for
validation) across $192$ translation directions. Table~\ref{tab:rl_data_full} reports
the number of sentences per direction, following the language-code convention of \citet{shang2026scalingmodeldatamultilingual}.

\begin{table*}[!ht]
\centering
\small
\setlength{\tabcolsep}{4pt}
\begin{tabular}{lrr | lrr | lrr}
\toprule
\textbf{Direction} & \textbf{Train} & \textbf{Val} &
\textbf{Direction} & \textbf{Train} & \textbf{Val} &
\textbf{Direction} & \textbf{Train} & \textbf{Val} \\
\hline
\hline
\texttt{en} $\rightarrow$ \texttt{ru} & 2205 & 73 & \texttt{cs} $\rightarrow$ \texttt{ta} & 71 & 2 & \texttt{hu} $\rightarrow$ \texttt{es} & 32 & 1 \\
\texttt{en} $\rightarrow$ \texttt{cs} & 1776 & 68 & \texttt{pl} $\rightarrow$ \texttt{yue} & 68 & 4 & \texttt{zhs} $\rightarrow$ \texttt{hi} & 32 & 1 \\
\texttt{en} $\rightarrow$ \texttt{de} & 1479 & 47 & \texttt{hu} $\rightarrow$ \texttt{th} & 68 & 1 & \texttt{ur} $\rightarrow$ \texttt{ru} & 31 & 2 \\
\texttt{en} $\rightarrow$ \texttt{ja} & 1327 & 43 & \texttt{ta} $\rightarrow$ \texttt{sl} & 63 & 0 & \texttt{ro} $\rightarrow$ \texttt{pt} & 31 & 1 \\
\texttt{en} $\rightarrow$ \texttt{bn} & 1219 & 38 & \texttt{bn} $\rightarrow$ \texttt{ar} & 62 & 0 & \texttt{uz} $\rightarrow$ \texttt{kk} & 31 & 0 \\
\texttt{ja} $\rightarrow$ \texttt{en} & 1189 & 47 & \texttt{ru} $\rightarrow$ \texttt{tr} & 60 & 0 & \texttt{nb} $\rightarrow$ \texttt{fa} & 30 & 2 \\
\texttt{en} $\rightarrow$ \texttt{zhs} & 775 & 18 & \texttt{my} $\rightarrow$ \texttt{en} & 57 & 3 & \texttt{bg} $\rightarrow$ \texttt{sk} & 30 & 1 \\
\bottomrule
\end{tabular}
\caption{Number of instances for each translation direction in the RL training
data, sorted by training count. All $192$ directions are shown ($30{,}572$ training /
$1{,}000$ validation in total).}
\label{tab:rl_data_full}
\end{table*}

\section{Language Details}\label{sec:appendix}
Table~\ref{tab:langs} provides information on the 46 languages supported by our models.

\begin{table*}[h]
\centering
\begin{tabular}{c | c | c | c | c | c}
ISO Code & Language & Script & Family & Subgrouping & Resource \\
\hline
\hline
\texttt{ar} & Arabic & Arabic & Afro-Asiatic & Semitic & High \\
\texttt{az} & Azerbaijani & Latin & Turkic & Common Turkic & Low\\
\texttt{bg} & Bulgarian & Cyrillic & Indo-European & Balto-Slavic & Mid \\
\texttt{bn} & Bengali & Bengali & Indo-European & Indo-Aryan & Mid \\
\texttt{ca} & Catalan & Latin & Indo-European & Italic & High \\
\texttt{cs} & Czech & Latin &  Indo-European & Balto-Slavic & High \\
\texttt{da} & Danish & Latin & Indo-European & Germanic & Mid \\
\hline
\end{tabular}
\caption{$46$ languages supported by our model. The resource of each language is determined according to the taxonomy classes by \citet{joshi-etal-2020-state}.}
\label{tab:langs}
\end{table*}

\section{Training Hyper-parameters}\label{app:hyperparams}
Table~\ref{tab:hyperparams} lists the complete set of GRPO hyper-parameters, shared across the $1$B, $4$B, and $12$B models.

\begin{table}[h]
\centering
\small
\begin{tabular}{ll}
\toprule
\textbf{Hyper-parameter} & \textbf{Value} \\
\midrule
\multicolumn{2}{l}{\textit{Optimization}} \\
Actor learning rate        & $1\mathrm{e}{-6}$ \\
Batch size (prompts)       & $128$ \\
PPO mini-batch size        & $128$ \\
Total epochs                     & $3$ \\
KL loss type               & low-variance KL \\
KL coefficient $\beta$     & $1\mathrm{e}{-3}$ \\
Gradient checkpointing     & Enabled \\
\midrule
\multicolumn{2}{l}{\textit{Rollout / sampling}} \\
Inference backend          & vLLM \\
Group size $G$             & $8$ \\
Temperature                & $0.8$ \\
Top-$p$                    & $0.95$ \\
Max prompt length          & $4096$ tokens \\
Max response length        & $4096$ tokens \\
\midrule
\multicolumn{2}{l}{\textit{Reward (reference-free)}} \\
Quality metrics            & XCOMET, COMETKiwi \\
Language-matched reward    & $\tfrac{1}{2}(\text{XCOMET}+\text{COMETKiwi})$ \\
Language-mismatch reward   & $0$ \\
Language identification    & OpenLID-v3 \\
\bottomrule
\end{tabular}
\caption{GRPO training hyper-parameters, shared across all three model scales.}
\label{tab:hyperparams}
\end{table}

\section{Detailed Evaluation Results}\label{app:detailed_results}
\subsection{Reference-Free Results}\label{app:reference_free_results}
Tables~\ref{tab:wmt24_baseline_en}--\ref{tab:flores_plus_ours_zh} report the per-direction reference-free results (XCOMET / COMETKiwi).

\begin{table*}[!ht]
\setlength{\tabcolsep}{2pt}
\centering
\resizebox{\textwidth}{!}{
    \begin{tabular}{c | c c c | c c | c c c c c c | c c | c c c}
Direction & TranslateGemma-4B & TranslateGemma-12B & TranslateGemma-27B & GemmaX2-28-2B & GemmaX2-28-9B & Hunyuan-MT-7B & HY-MT1.5-1.8B & HY-MT1.5-7B & HY-MT2-1.8B & HY-MT2-7B & HY-MT2-30B-A3B & Seed-X-Instruct-7B & Seed-X-PPO-7B & Tower-Plus-2B & Tower-Plus-9B & Tower-Plus-72B \\
\hline
\hline
    \texttt{en} $\rightarrow$ \texttt{ar} & 82.93 / 78.21 & 86.54 / 81.66 & 87.12 / 83.01 & 77.96 / 74.33 & 82.34 / 78.80 & 88.87 / 82.85 & 85.12 / 78.51 & 88.98 / 82.66 & 83.57 / 78.39 & 87.70 / 82.24 & 88.27 / 82.49 & 80.33 / 75.78 & 81.80 / 77.34 & - / - & - / - & - / - \\
    \hline
    \texttt{en} $\rightarrow$ \texttt{az} & 39.43 / 45.81 & 71.65 / 76.53 & 77.85 / 81.72 & - / - & - / - & - / - & - / - & - / - & - / - & - / - & - / - & - / - & - / - & - / - & - / - & - / - \\
    \hline
    \texttt{en} $\rightarrow$ \texttt{bg} & 83.19 / 77.11 & 88.78 / 83.66 & 89.77 / 84.98 & - / - & - / - & - / - & - / - & - / - & - / - & - / - & - / - & - / - & - / - & - / - & - / - & - / - \\
    \hline
    \texttt{en} $\rightarrow$ \texttt{bn} & 78.79 / 75.57 & 84.04 / 79.68 & 84.95 / 80.72 & 78.80 / 77.55 & 81.61 / 79.78 & 86.50 / 80.30 & 83.05 / 75.73 & 86.77 / 80.81 & 81.93 / 77.34 & 86.77 / 81.94 & 87.50 / 82.22 & - / - & - / - & - / - & - / - & - / - \\
    \hline
    \texttt{en} $\rightarrow$ \texttt{ca} & 79.59 / 71.97 & 85.65 / 78.91 & 86.59 / 80.12 & - / - & - / - & - / - & - / - & - / - & - / - & - / - & - / - & - / - & - / - & - / - & - / - & - / - \\
    \hline
    \texttt{en} $\rightarrow$ \texttt{cs} & 80.07 / 75.12 & 86.54 / 81.74 & 87.55 / 83.54 & 78.78 / 71.86 & 83.84 / 77.78 & 88.05 / 82.53 & 82.61 / 75.88 & 87.58 / 82.28 & 82.60 / 74.60 & 87.49 / 80.68 & 88.50 / 82.13 & 83.43 / 76.69 & 84.35 / 77.67 & 76.91 / 69.83 & 83.61 / 77.82 & 83.06 / 76.78 \\
\end{tabular}}
\caption{Reference-free evaluation results (XCOMET / COMETKiwi) of baseline models (Part I) on the WMT24++ benchmark.}\label{tab:wmt24_baseline_en}
\end{table*}

\begin{table*}[!ht]
\setlength{\tabcolsep}{2pt}
\centering
\resizebox{\textwidth}{!}{
    \begin{tabular}{c | c c c c c | c c c c c c}
Direction & Google Translate & NLLB-54.5B & Gemini 2.5 Pro & Gemini 3 Pro & GPT-5 & MiLMMT-46-1B-v0.1 & MiLMMT-46-1B-v1.0 & MiLMMT-46-4B-v0.1 & MiLMMT-46-4B-v1.0 & MiLMMT-46-12B-v0.1 & MiLMMT-46-12B-v1.0 \\
\hline
\hline
    \texttt{en} $\rightarrow$ \texttt{ar} & 86.44 / 81.97 & - / - & 84.85 / 81.08 & 85.10 / 81.20 & 84.36 / 81.56 & 74.32 / 71.17 & 78.72 / 75.15 & 82.17 / 78.59 & 86.67 / 81.80 & 83.68 / 80.37 & 89.44 / 84.33 \\
    \hline
    \texttt{en} $\rightarrow$ \texttt{az} & 75.04 / 78.02 & - / - & 78.83 / 81.72 & 80.01 / 82.17 & 80.36 / 82.29 & 66.15 / 69.98 & 73.65 / 75.98 & 77.81 / 79.42 & 83.99 / 83.98 & 80.36 / 81.29 & 87.76 / 86.56 \\
    \hline
    \texttt{en} $\rightarrow$ \texttt{bg} & 85.07 / 80.04 & - / - & 86.98 / 81.51 & 87.13 / 81.84 & 87.24 / 82.49 & 78.31 / 72.59 & 83.44 / 77.26 & 85.18 / 79.44 & 89.97 / 84.08 & 87.18 / 81.96 & 91.83 / 86.83 \\
    \hline
    \texttt{en} $\rightarrow$ \texttt{bn} & 76.65 / 76.64 & - / - & 81.98 / 79.63 & 84.39 / 81.09 & 82.14 / 80.01 & 73.35 / 72.87 & 80.44 / 77.38 & 81.13 / 79.21 & 87.39 / 81.83 & 82.97 / 80.67 & 89.00 / 83.30 \\
    \hline
    \texttt{en} $\rightarrow$ \texttt{ca} & 82.82 / 76.02 & - / - & 84.16 / 77.08 & 84.97 / 77.73 & 85.39 / 78.61 & 79.25 / 70.51 & 82.62 / 74.97 & 84.88 / 77.31 & 88.83 / 81.39 & 85.82 / 78.49 & 90.60 / 83.50 \\
    \hline
    \texttt{en} $\rightarrow$ \texttt{cs} & 83.42 / 77.94 & - / - & 85.82 / 80.02 & 85.95 / 80.36 & 86.08 / 81.57 & 74.07 / 66.96 & 79.72 / 73.77 & 82.84 / 77.04 & 87.89 / 82.89 & 85.99 / 80.36 & 90.08 / 85.44 \\
    \hline
    \texttt{en} $\rightarrow$ \texttt{da} & 90.57 / 81.72 & - / - & 91.88 / 83.64 & 92.15 / 84.16 & 92.06 / 84.59 & 86.69 / 76.36 & 89.50 / 79.45 & 90.86 / 81.85 & 93.55 / 85.42 & 92.14 / 83.36 & 95.05 / 87.05 \\
\end{tabular}}
\caption{Reference-free evaluation results (XCOMET / COMETKiwi) of baseline models (Part II) and MiLMMT models on the WMT24++ benchmark.}\label{tab:wmt24_ours_en}
\end{table*}

\begin{table*}[!ht]
\setlength{\tabcolsep}{2pt}
\centering
\resizebox{\textwidth}{!}{
    \begin{tabular}{c | c c c | c c | c c c c c c | c c | c c c}
Direction & TranslateGemma-4B & TranslateGemma-12B & TranslateGemma-27B & GemmaX2-28-2B & GemmaX2-28-9B & Hunyuan-MT-7B & HY-MT1.5-1.8B & HY-MT1.5-7B & HY-MT2-1.8B & HY-MT2-7B & HY-MT2-30B-A3B & Seed-X-Instruct-7B & Seed-X-PPO-7B & Tower-Plus-2B & Tower-Plus-9B & Tower-Plus-72B \\
\hline
\hline
    \texttt{en} $\rightarrow$ \texttt{ar} & 93.54 / 87.09 & 95.66 / 89.84 & 96.38 / 90.84 & 91.90 / 85.63 & 94.20 / 87.97 & 96.24 / 90.07 & 93.88 / 86.84 & 96.05 / 90.22 & 93.32 / 86.88 & 95.60 / 89.57 & 95.70 / 89.69 & 93.34 / 87.04 & 93.78 / 87.76 & - / - & - / - & - / - \\
    \texttt{ar} $\rightarrow$ \texttt{en} & 87.46 / 84.55 & 90.50 / 85.89 & 91.93 / 86.59 & 91.31 / 86.44 & 92.65 / 87.19 & 90.88 / 85.58 & 88.41 / 83.29 & 91.20 / 85.75 & 89.88 / 85.05 & 92.30 / 86.81 & 92.62 / 86.69 & 90.08 / 85.88 & 92.70 / 86.99 & - / - & - / - & - / - \\
    \hline
    \texttt{en} $\rightarrow$ \texttt{az} & 50.49 / 54.71 & 87.63 / 87.30 & 92.72 / 91.02 & - / - & - / - & - / - & - / - & - / - & - / - & - / - & - / - & - / - & - / - & - / - & - / - & - / - \\
    \texttt{az} $\rightarrow$ \texttt{en} & 67.08 / 83.64 & 76.62 / 86.34 & 81.93 / 87.87 & - / - & - / - & - / - & - / - & - / - & - / - & - / - & - / - & - / - & - / - & - / - & - / - & - / - \\
    \hline
    \texttt{en} $\rightarrow$ \texttt{bg} & 94.35 / 89.46 & 97.23 / 93.44 & 97.88 / 94.36 & - / - & - / - & - / - & - / - & - / - & - / - & - / - & - / - & - / - & - / - & - / - & - / - & - / - \\
    \texttt{bg} $\rightarrow$ \texttt{en} & 91.52 / 89.06 & 93.40 / 90.01 & 94.93 / 90.79 & - / - & - / - & - / - & - / - & - / - & - / - & - / - & - / - & - / - & - / - & - / - & - / - & - / - \\
    \hline
    \texttt{en} $\rightarrow$ \texttt{bn} & 90.38 / 81.97 & 93.58 / 85.14 & 94.19 / 85.99 & 91.40 / 84.35 & 93.15 / 86.05 & 94.81 / 86.03 & 92.21 / 81.61 & 95.05 / 86.60 & 91.99 / 82.94 & 94.80 / 86.89 & 95.37 / 87.59 & - / - & - / - & - / - & - / - & - / - \\
\end{tabular}}
\caption{Reference-free English-centric evaluation results (XCOMET / COMETKiwi) of baseline models (Part I) on the FLORES+ benchmark.}\label{tab:flores_plus_baseline_en}
\end{table*}

\begin{table*}[!ht]
\setlength{\tabcolsep}{2pt}
\centering
\resizebox{\textwidth}{!}{
    \begin{tabular}{c | c c c c c | c c c c c c}
Direction & Google Translate & NLLB-54.5B & Gemini 2.5 Pro & Gemini 3 Pro & GPT-5 & MiLMMT-46-1B-v0.1 & MiLMMT-46-1B-v1.0 & MiLMMT-46-4B-v0.1 & MiLMMT-46-4B-v1.0 & MiLMMT-46-12B-v0.1 & MiLMMT-46-12B-v1.0 \\
\hline
\hline
    \texttt{en} $\rightarrow$ \texttt{ar} & 95.05 / 89.48 & 91.61 / 84.52 & 95.64 / 89.85 & 95.68 / 90.03 & 94.83 / 89.87 & 89.03 / 82.35 & 91.63 / 85.10 & 93.67 / 87.88 & 95.49 / 89.82 & 94.77 / 88.90 & 96.96 / 91.53 \\
    \texttt{ar} $\rightarrow$ \texttt{en} & 92.15 / 87.00 & 90.63 / 85.83 & 92.75 / 87.02 & 92.56 / 87.07 & 92.00 / 86.93 & 90.24 / 85.76 & 91.34 / 86.04 & 92.39 / 86.93 & 92.66 / 87.06 & 93.11 / 87.28 & 93.35 / 87.37 \\
    \hline
    \texttt{en} $\rightarrow$ \texttt{az} & 88.43 / 86.98 & 85.93 / 82.51 & 93.82 / 92.04 & 94.15 / 92.24 & 94.28 / 92.14 & 84.73 / 83.88 & 88.48 / 87.01 & 92.15 / 90.09 & 94.86 / 92.54 & 93.63 / 91.42 & 96.67 / 93.96 \\
    \texttt{az} $\rightarrow$ \texttt{en} & 85.50 / 88.87 & 81.21 / 87.65 & 86.56 / 88.83 & 86.65 / 88.77 & 86.08 / 88.74 & 80.83 / 86.51 & 83.97 / 86.94 & 86.01 / 88.53 & 87.50 / 88.38 & 86.60 / 88.84 & 88.19 / 88.80 \\
    \hline
    \texttt{en} $\rightarrow$ \texttt{bg} & 95.99 / 91.09 & 93.16 / 86.35 & 96.79 / 93.02 & 97.26 / 93.28 & 96.10 / 93.18 & 92.30 / 86.17 & 94.08 / 88.79 & 96.02 / 91.61 & 97.41 / 94.00 & 96.73 / 92.41 & 98.28 / 95.08 \\
    \texttt{bg} $\rightarrow$ \texttt{en} & 95.26 / 91.09 & 92.30 / 89.50 & 95.36 / 91.09 & 95.47 / 91.16 & 94.87 / 91.10 & 94.17 / 90.26 & 94.80 / 90.39 & 95.64 / 91.17 & 96.00 / 91.33 & 95.70 / 91.37 & 95.89 / 91.39 \\
    \hline
    \texttt{en} $\rightarrow$ \texttt{bn} & 90.40 / 84.17 & 87.81 / 80.98 & 92.82 / 85.51 & 94.31 / 86.87 & 91.56 / 85.61 & 86.95 / 81.05 & 91.83 / 83.94 & 92.52 / 85.61 & 95.51 / 87.45 & 93.64 / 86.36 & 96.69 / 88.49 \\
\end{tabular}}
\caption{Reference-free English-centric evaluation results (XCOMET / COMETKiwi) of baseline models (Part II) and MiLMMT models on the FLORES+ benchmark.}\label{tab:flores_plus_ours_en}
\end{table*}

\begin{table*}[!ht]
\setlength{\tabcolsep}{2pt}
\centering
\resizebox{\textwidth}{!}{
    \begin{tabular}{c | c c c | c c | c c c c c c | c c | c c c}
Direction & TranslateGemma-4B & TranslateGemma-12B & TranslateGemma-27B & GemmaX2-28-2B & GemmaX2-28-9B & Hunyuan-MT-7B & HY-MT1.5-1.8B & HY-MT1.5-7B & HY-MT2-1.8B & HY-MT2-7B & HY-MT2-30B-A3B & Seed-X-Instruct-7B & Seed-X-PPO-7B & Tower-Plus-2B & Tower-Plus-9B & Tower-Plus-72B \\
\hline
\hline
    \texttt{zhs} $\rightarrow$ \texttt{ar} & 78.48 / 73.15 & 81.62 / 76.79 & 83.13 / 77.26 & 77.98 / 71.25 & 82.40 / 74.60 & 83.67 / 76.35 & 80.80 / 73.82 & 84.28 / 77.25 & 81.46 / 73.69 & 83.97 / 76.58 & 83.39 / 75.80 & 78.17 / 71.29 & 82.11 / 75.08 & - / - & - / - & - / - \\
    \texttt{ar} $\rightarrow$ \texttt{zhs} & 68.08 / 68.66 & 70.55 / 70.69 & 71.60 / 71.19 & 68.83 / 70.57 & 71.75 / 72.40 & 73.04 / 71.40 & 70.80 / 69.76 & 73.87 / 71.00 & 72.49 / 70.83 & 74.49 / 71.46 & 74.36 / 72.82 & 65.91 / 70.45 & 65.42 / 69.60 & - / - & - / - & - / - \\
    \hline
    \texttt{zhs} $\rightarrow$ \texttt{az} & 40.12 / 51.10 & 75.66 / 81.55 & 81.98 / 84.82 & - / - & - / - & - / - & - / - & - / - & - / - & - / - & - / - & - / - & - / - & - / - & - / - & - / - \\
    \texttt{az} $\rightarrow$ \texttt{zhs} & 42.27 / 73.69 & 49.12 / 78.89 & 50.71 / 79.92 & - / - & - / - & - / - & - / - & - / - & - / - & - / - & - / - & - / - & - / - & - / - & - / - & - / - \\
    \hline
    \texttt{zhs} $\rightarrow$ \texttt{bg} & 88.55 / 81.80 & 92.12 / 85.72 & 92.94 / 86.44 & - / - & - / - & - / - & - / - & - / - & - / - & - / - & - / - & - / - & - / - & - / - & - / - & - / - \\
    \texttt{bg} $\rightarrow$ \texttt{zhs} & 66.45 / 73.69 & 68.36 / 76.21 & 69.48 / 76.60 & - / - & - / - & - / - & - / - & - / - & - / - & - / - & - / - & - / - & - / - & - / - & - / - & - / - \\
    \hline
    \texttt{zhs} $\rightarrow$ \texttt{bn} & 84.89 / 72.44 & 87.52 / 76.05 & 89.18 / 76.30 & 84.72 / 73.63 & 88.10 / 75.71 & 87.97 / 74.38 & 84.46 / 71.06 & 89.20 / 77.23 & 85.64 / 72.10 & 90.41 / 75.98 & 90.58 / 76.10 & - / - & - / - & - / - & - / - & - / - \\
\end{tabular}}
\caption{Reference-free Chinese-centric evaluation results (XCOMET / COMETKiwi) of baseline models (Part I) on the FLORES+ benchmark.}\label{tab:flores_plus_baselines_zh}
\end{table*}

\begin{table*}[!ht]
\setlength{\tabcolsep}{2pt}
\centering
\resizebox{\textwidth}{!}{
    \begin{tabular}{c | c c c c c | c c c c c c}
Direction & Google Translate & NLLB-54.5B & Gemini 2.5 Pro & Gemini 3 Pro & GPT-5 & MiLMMT-46-1B-v0.1 & MiLMMT-46-1B-v1.0 & MiLMMT-46-4B-v0.1 & MiLMMT-46-4B-v1.0 & MiLMMT-46-12B-v0.1 & MiLMMT-46-12B-v1.0 \\
\hline
\hline
    \texttt{zhs} $\rightarrow$ \texttt{ar} & 79.53 / 73.09 & 79.30 / 71.84 & 82.51 / 74.67 & 83.14 / 74.76 & 83.06 / 75.46 & 74.88 / 68.37 & 77.85 / 71.85 & 81.05 / 73.92 & 83.96 / 76.94 & 83.03 / 75.40 & 84.86 / 78.05 \\
    \texttt{ar} $\rightarrow$ \texttt{zhs} & 66.40 / 70.69 & 52.06 / 58.70 & 70.50 / 72.42 & 71.81 / 72.89 & 71.76 / 73.05 & 67.78 / 68.55 & 71.52 / 70.30 & 70.30 / 71.67 & 73.99 / 72.37 & 71.87 / 73.29 & 75.00 / 73.59 \\
    \hline
    \texttt{zhs} $\rightarrow$ \texttt{az} & 76.37 / 79.72 & 74.45 / 76.04 & 84.65 / 85.37 & 84.75 / 85.69 & 85.47 / 86.09 & 72.42 / 77.31 & 76.26 / 80.69 & 82.79 / 84.23 & 84.44 / 86.17 & 84.49 / 85.74 & 86.96 / 87.57 \\
    \texttt{az} $\rightarrow$ \texttt{zhs} & 42.77 / 79.66 & 35.20 / 69.50 & 51.11 / 81.48 & 51.15 / 82.01 & 53.65 / 82.06 & 44.88 / 75.64 & 50.64 / 77.61 & 49.20 / 80.64 & 54.71 / 80.90 & 50.47 / 81.51 & 57.09 / 81.96 \\
    \hline
    \texttt{zhs} $\rightarrow$ \texttt{bg} & 90.64 / 82.96 & 87.00 / 78.52 & 92.39 / 85.37 & 92.66 / 85.35 & 93.18 / 86.06 & 86.60 / 78.50 & 88.93 / 81.56 & 92.11 / 84.75 & 93.20 / 86.45 & 92.86 / 85.62 & 93.85 / 87.51 \\
    \texttt{bg} $\rightarrow$ \texttt{zhs} & 63.15 / 73.91 & 41.99 / 52.23 & 67.76 / 76.00 & 67.38 / 75.83 & 68.75 / 76.29 & 64.10 / 72.59 & 69.84 / 74.82 & 67.40 / 75.25 & 72.61 / 77.68 & 68.25 / 76.39 & 73.58 / 78.85 \\
    \hline
    \texttt{zhs} $\rightarrow$ \texttt{bn} & 83.30 / 70.64 & 80.46 / 67.99 & 87.97 / 75.47 & 88.97 / 76.23 & 89.13 / 76.45 & 79.75 / 69.11 & 85.07 / 74.04 & 87.50 / 75.32 & 90.12 / 78.07 & 88.52 / 76.34 & 91.34 / 78.99 \\
\end{tabular}}
\caption{Reference-free Chinese-centric evaluation results (XCOMET / COMETKiwi) of baseline models (Part II) and MiLMMT models on the FLORES+ benchmark.}\label{tab:flores_plus_ours_zh}
\end{table*}

\input{appendix_reference_tables.tex}

\section{Full On-Policy Distillation Results}\label{app:opd_full}

\paragraph{OPD Training Details.}
We adopt optimization settings similar to those used for GRPO in our
OPD experiments, including a learning rate of $1\times10^{-6}$, a
prompt batch size of $128$, a PPO mini-batch size of $128$, three
training epochs, and maximum prompt and response lengths of $4096$
tokens each. For RL+OPD, we use the same task reward and group size
$G=8$ as in GRPO. Pure OPD does not use the task reward and instead
samples one student trajectory per prompt. We use the fixed
MiLMMT-46-12B-v1.0 model as the teacher and the $k_1$ estimator for
PG-OPD. For the GKD control, we use a top-$128$ approximation to the
forward KL divergence.

\paragraph{Full Results.}
Table~\ref{tab:opd_full} reports the complete OPD and RL+OPD results
underlying Section~\ref{sec:opd} and Table~\ref{tab:opd}, as $46$-language
macro-averages. All students are distilled from the fixed MiLMMT-46-12B-v1.0
teacher. \emph{RL+OPD} adds the QE-based task reward with distillation weight
$\lambda$; \emph{OPD} rows use only the distillation loss. Unless noted, the
student is initialized from v0.1; \emph{init.\ v1.0} rows are initialized from the interpolated
checkpoint instead. WMT24++ cells report reference-free XCOMET / COMETKiwi;
each FLORES+ direction group reports spBLEU / reference-based XCOMET /
reference-free XCOMET / COMETKiwi.

\begin{table*}[!ht]
\centering
\small
\setlength{\tabcolsep}{2pt}
\resizebox{\textwidth}{!}{%
\begin{tabular}{l | c | c c c c}
\hline
\multicolumn{1}{c|}{Model} & WMT24++ & \multicolumn{4}{c}{FLORES+} \\
& \texttt{en} $\rightarrow$ \texttt{xx} & \texttt{en} $\rightarrow$ \texttt{xx} & \texttt{xx} $\rightarrow$ \texttt{en} & \texttt{zh} $\rightarrow$ \texttt{xx} & \texttt{xx} $\rightarrow$ \texttt{zh} \\
\hline
\hline
\textbf{$1$B students} & & & & & \\
RL+OPD $\lambda{=}1$ & 78.25/75.35 & 33.87/87.79/90.71/86.56 & 39.65/91.17/90.65/88.27 & 21.64/82.24/82.76/77.77 & 26.77/81.48/67.18/76.93 \\
RL+OPD $\lambda{=}0.5$ & 78.58/75.69 & 33.76/87.90/90.90/86.72 & 39.41/91.24/90.75/88.30 & 21.63/82.43/82.97/78.00 & 26.70/81.57/67.50/77.10 \\
RL+OPD $\lambda{=}0.1$ & 79.37/76.41 & 33.32/88.38/91.39/87.19 & 38.56/91.44/91.02/88.36 & 21.34/82.93/83.66/78.61 & 26.40/81.89/68.14/77.43 \\
RL+OPD $\lambda{=}0.05$ & 79.83/76.68 & 32.66/88.49/91.56/87.29 & 38.02/91.47/91.11/88.37 & 21.04/83.00/83.72/78.80 & 26.04/82.09/68.39/77.84 \\
RL+OPD $\lambda{=}0.01$ & 80.29/77.01 & 31.11/88.53/91.74/87.29 & 37.33/91.45/91.14/88.33 & 20.47/83.12/83.95/78.97 & 24.89/82.10/68.74/77.78 \\
RL+OPD $\lambda{=}0.001$ & 80.42/77.08 & 29.65/88.37/91.66/87.29 & 36.30/91.46/91.15/88.24 & 19.70/82.82/83.82/79.14 & 24.12/82.05/68.51/77.79 \\
RL+OPD $\lambda{=}10^{-4}$ & 80.26/76.92 & 30.03/88.32/91.62/87.19 & 36.65/91.45/91.11/88.27 & 19.89/82.82/83.93/79.08 & 24.18/82.04/69.08/77.97 \\
OPD (PG) & 77.67/74.66 & 33.95/87.19/90.21/86.03 & 39.78/90.97/90.42/88.19 & 21.44/81.63/82.13/77.17 & 26.70/81.07/66.70/76.58 \\
OPD (GKD) & 77.42/74.37 & 32.90/87.12/90.11/85.91 & 39.54/90.96/90.44/88.15 & 20.83/81.32/81.77/76.86 & 26.35/81.02/66.56/76.40 \\
OPD (PG, init. v1.0) & 77.86/74.93 & 33.87/87.45/90.43/86.29 & 39.58/91.06/90.54/88.23 & 21.43/81.93/82.40/77.42 & 26.69/81.23/66.87/76.68 \\
\hline
\hline
\textbf{$4$B students} & & & & & \\
RL+OPD $\lambda{=}1$ & 85.85/82.23 & 37.09/93.34/95.15/91.12 & 42.70/93.98/92.78/89.41 & 25.44/89.92/88.56/82.36 & 30.68/86.67/70.43/79.99 \\
RL+OPD $\lambda{=}0.5$ & 86.03/82.45 & 36.81/93.44/95.27/91.27 & 42.43/94.03/92.86/89.42 & 25.22/89.96/88.60/82.51 & 30.43/86.89/70.59/80.17 \\
RL+OPD $\lambda{=}0.1$ & 86.67/82.95 & 35.40/93.72/95.57/91.57 & 41.59/94.11/93.00/89.42 & 24.55/90.13/88.93/82.85 & 29.67/87.27/71.50/80.57 \\
RL+OPD $\lambda{=}0.05$ & 87.03/83.17 & 34.25/93.74/95.62/91.66 & 41.22/94.10/93.03/89.41 & 24.01/90.09/88.97/82.95 & 29.18/87.30/71.36/80.55 \\
RL+OPD $\lambda{=}0.01$ & 87.62/83.46 & 31.76/93.89/95.88/91.77 & 41.06/94.16/92.99/89.39 & 23.38/90.19/88.96/83.28 & 28.19/87.61/71.41/80.66 \\
RL+OPD $\lambda{=}0.001$ & 87.68/83.57 & 30.35/93.87/95.88/91.80 & 40.64/94.13/92.95/89.38 & 22.70/90.16/89.03/83.45 & 27.12/87.71/71.98/80.94 \\
RL+OPD $\lambda{=}10^{-4}$ & 87.74/83.45 & 30.41/93.85/95.83/91.74 & 40.27/94.14/92.99/89.36 & 22.63/90.06/89.01/83.52 & 26.63/87.63/71.56/80.67 \\
OPD (PG) & 85.42/81.86 & 37.56/93.15/94.97/90.94 & 42.88/93.91/92.68/89.37 & 25.46/89.74/88.36/82.13 & 30.80/86.47/70.01/79.79 \\
OPD (GKD) & 85.62/81.96 & 36.97/93.21/95.01/91.01 & 42.68/93.94/92.76/89.38 & 25.28/89.76/88.36/82.25 & 30.71/86.58/70.08/79.88 \\
OPD (PG, init. v1.0) & 85.37/81.85 & 37.53/93.14/94.97/90.97 & 42.78/93.89/92.71/89.40 & 25.47/89.69/88.32/82.09 & 30.85/86.45/70.08/79.76 \\
\hline
\end{tabular}%
}
\caption{Full on-policy distillation results as $46$-language. All students are distilled from the fixed MiLMMT-46-12B-v1.0 teacher. WMT24++ cells report reference-free XCOMET / COMETKiwi; each FLORES+ cell reports spBLEU / reference-based XCOMET / reference-free XCOMET / COMETKiwi.}
\label{tab:opd_full}
\end{table*}

\end{document}

%% file: main_results_table.tex
\begin{table*}[!ht]
\centering
\small
\setlength{\tabcolsep}{2pt}
\resizebox{\textwidth}{!}{%
\begin{tabular}{l | c | cc | cc | cc | cc}
\hline
\multicolumn{1}{c|}{Model} & \multicolumn{1}{c|}{WMT24++} & \multicolumn{8}{c}{FLORES+} \\
& \texttt{en} $\rightarrow$ \texttt{xx} & \multicolumn{2}{c|}{\texttt{en} $\rightarrow$ \texttt{xx}} & \multicolumn{2}{c|}{\texttt{xx} $\rightarrow$ \texttt{en}} & \multicolumn{2}{c|}{\texttt{zh} $\rightarrow$ \texttt{xx}} & \multicolumn{2}{c}{\texttt{xx} $\rightarrow$ \texttt{zh}} \\
& QE & Ref. & QE & Ref. & QE & Ref. & QE & Ref. & QE \\
\hline
\hline
\textbf{$21$ languages} & & & & & & & & & \\
Tower-Plus-2B & 82.52/76.30 & 40.41/92.65 & 94.10/88.53 & 42.54/94.61 & 94.88/89.77 & 24.27/88.56 & 87.99/78.87 & 29.53/84.65 & 71.29/76.52 \\
Tower-Plus-9B & 86.80/80.76 & \textbf{43.33}/95.09 & 95.93/91.05 & \underline{45.32}/96.03 & 95.84/90.44 & 27.59/92.65 & 90.73/80.68 & \underline{33.25}/88.09 & 73.95/78.59 \\
Tower-Plus-72B & 86.25/80.21 & \underline{42.74}/94.86 & 95.76/90.71 & \textbf{45.89}/\underline{96.26} & 96.00/90.51 & \textbf{28.50}/\underline{92.92} & 90.55/80.41 & \textbf{35.06}/89.17 & 74.61/78.88 \\
MiLMMT-46-1B-v1.0 & 83.28/76.69 & 36.87/92.02 & 93.88/88.03 & 41.10/94.45 & 95.15/89.73 & 22.94/87.49 & 87.57/78.29 & 28.20/85.39 & 74.03/77.78 \\
MiLMMT-46-4B-v1.0 & \underline{89.11}/\underline{82.77} & 39.85/\underline{95.64} & \underline{96.67}/\underline{91.85} & 43.45/96.06 & \underline{96.33}/\underline{90.57} & 26.53/92.83 & \underline{91.38}/\underline{82.07} & 31.47/\underline{89.49} & \underline{76.77}/\underline{80.45} \\
MiLMMT-46-12B-v1.0 & \textbf{90.77}/\textbf{84.55} & 40.66/\textbf{96.60} & \textbf{97.39}/\textbf{92.88} & 44.59/\textbf{96.55} & \textbf{96.53}/\textbf{90.77} & \underline{28.03}/\textbf{94.43} & \textbf{92.28}/\textbf{83.14} & 32.19/\textbf{90.79} & \textbf{78.01}/\textbf{81.43} \\
\hline
\hline
\textbf{$26$ languages} & & & & & & & & & \\
Seed-X-Instruct-7B & 85.19/78.30 & \underline{44.16}/94.45 & 95.30/90.17 & \underline{44.54}/93.10 & 91.92/88.67 & \underline{28.60}/90.43 & 85.68/80.96 & \textbf{32.51}/87.60 & 68.36/79.97 \\
Seed-X-PPO-7B & 86.18/79.99 & \textbf{45.48}/95.02 & 95.95/91.10 & 44.12/95.92 & 95.73/90.51 & \textbf{29.65}/\underline{92.80} & 89.32/83.88 & 29.20/88.86 & 68.90/80.88 \\
MiLMMT-46-1B-v1.0 & 83.42/77.21 & 37.72/92.39 & 94.19/88.66 & 41.72/94.28 & 94.87/89.82 & 23.12/87.32 & 86.75/80.83 & 27.89/85.04 & 73.44/79.45 \\
MiLMMT-46-4B-v1.0 & \underline{89.19}/\underline{83.25} & 40.50/\underline{95.87} & \underline{96.86}/\underline{92.40} & 44.30/\underline{95.94} & \underline{96.06}/\underline{90.68} & 26.72/92.75 & \underline{90.63}/\underline{84.79} & 31.36/\underline{89.32} & \underline{76.30}/\underline{82.31} \\
MiLMMT-46-12B-v1.0 & \textbf{90.88}/\textbf{85.00} & 41.18/\textbf{96.84} & \textbf{97.58}/\textbf{93.46} & \textbf{45.43}/\textbf{96.45} & \textbf{96.28}/\textbf{90.89} & 28.15/\textbf{94.34} & \textbf{91.58}/\textbf{85.74} & \underline{32.39}/\textbf{90.72} & \textbf{77.58}/\textbf{83.28} \\
\hline
\hline
\textbf{$28$ languages} & & & & & & & & & \\
GemmaX2-28-2B-v0.2 & 80.95/77.35 & \underline{37.38}/90.14 & 92.42/87.69 & 42.29/92.52 & 91.86/88.28 & 25.06/86.56 & 85.32/79.71 & 29.70/83.57 & 68.60/78.17 \\
GemmaX2-28-9B-v0.2 & 84.19/80.28 & \textbf{40.04}/92.66 & 94.32/89.56 & \textbf{44.53}/\underline{94.02} & 92.88/\underline{88.81} & \textbf{27.67}/\underline{90.27} & 87.94/81.75 & \textbf{32.39}/86.42 & 70.41/79.78 \\
MiLMMT-46-1B-v1.0 & 79.73/75.97 & 33.12/88.23 & 91.22/86.37 & 38.73/91.24 & 91.23/87.73 & 21.43/83.24 & 83.36/78.35 & 26.28/81.48 & 68.84/77.60 \\
MiLMMT-46-4B-v1.0 & \underline{86.13}/\underline{81.81} & 36.16/\underline{93.19} & \underline{95.05}/\underline{90.30} & 42.05/93.86 & \underline{93.09}/88.75 & 25.01/90.03 & \underline{88.47}/\underline{82.29} & 30.11/\underline{86.55} & \underline{71.54}/\underline{80.32} \\
MiLMMT-46-12B-v1.0 & \textbf{88.09}/\textbf{83.57} & 36.82/\textbf{94.58} & \textbf{96.01}/\textbf{91.39} & \underline{43.44}/\textbf{94.62} & \textbf{93.52}/\textbf{89.00} & \underline{26.39}/\textbf{91.93} & \textbf{89.55}/\textbf{83.19} & \underline{31.35}/\textbf{88.20} & \textbf{72.84}/\textbf{81.30} \\
\hline
\hline
\textbf{$31$ languages} & & & & & & & & & \\
HY-MT1.5-1.8B & 83.88/77.38 & 24.81/88.51 & 91.80/85.31 & 25.25/88.51 & 87.51/84.01 & 17.60/82.79 & 84.93/74.74 & 22.61/82.58 & 70.06/77.06 \\
Hunyuan-MT-7B & 84.51/80.71 & 27.84/89.97 & 92.37/88.53 & 30.20/89.90 & 88.29/85.76 & 20.83/85.89 & 84.55/78.08 & 22.98/85.35 & 71.69/79.15 \\
HY-MT1.5-7B & 84.42/80.85 & 28.97/90.16 & 92.37/88.78 & 31.68/90.44 & 88.87/86.20 & 20.61/86.38 & 85.45/79.35 & 23.59/86.18 & 72.45/79.10 \\
HY-MT2-1.8B & 82.61/78.30 & 30.97/88.40 & 91.14/86.20 & 32.29/88.57 & 88.29/85.83 & 21.31/85.17 & 84.87/76.82 & 24.82/82.60 & 70.89/75.74 \\
HY-MT2-7B & 86.21/81.86 & 34.84/92.10 & 93.92/89.68 & 38.79/92.57 & 91.36/87.78 & 23.24/89.29 & 87.69/80.14 & 25.35/88.08 & \textbf{73.79}/78.36 \\
HY-MT2-30B-A3B & \underline{86.44}/\underline{82.58} & 34.94/92.30 & 93.87/90.24 & 40.07/\underline{93.90} & 91.98/88.08 & 24.38/\underline{89.92} & 87.79/78.25 & 29.91/\textbf{88.83} & \underline{73.68}/78.60 \\
MiLMMT-46-1B-v1.0 & 78.95/75.94 & 32.54/87.82 & 90.76/86.40 & 37.95/91.06 & 90.80/87.37 & 21.89/83.36 & 83.74/77.14 & 26.64/81.85 & 69.44/76.82 \\
MiLMMT-46-4B-v1.0 & 85.48/81.82 & \underline{35.61}/\underline{92.90} & \underline{94.71}/\underline{90.28} & \underline{41.24}/93.75 & \underline{92.69}/\underline{88.39} & \underline{25.30}/89.85 & \underline{88.66}/\underline{80.74} & \underline{30.36}/86.79 & 72.12/\underline{79.31} \\
MiLMMT-46-12B-v1.0 & \textbf{87.50}/\textbf{83.59} & \textbf{36.27}/\textbf{94.33} & \textbf{95.73}/\textbf{91.35} & \textbf{42.68}/\textbf{94.58} & \textbf{93.18}/\textbf{88.65} & \textbf{26.61}/\textbf{91.81} & \textbf{89.76}/\textbf{81.83} & \textbf{31.44}/\underline{88.38} & 73.35/\textbf{80.30} \\
\hline
\hline
\textbf{$46$ languages} & & & & & & & & & \\
Google Translate & 83.29/80.46 & \textbf{42.90}/91.39 & 92.71/88.97 & \textbf{47.42}/93.95 & 91.94/89.24 & \textbf{30.74}/89.08 & 83.94/78.26 & \textbf{36.08}/85.35 & 63.80/76.48 \\
Gemini 3 Pro & 85.03/81.70 & \underline{42.42}/\underline{94.17} & 95.13/91.20 & \underline{46.44}/\textbf{94.87} & 92.50/89.35 & \underline{29.90}/\textbf{92.22} & 88.49/81.66 & \underline{33.81}/87.45 & 68.48/78.49 \\
Gemini 2.5 Pro & 84.49/81.46 & 41.15/93.85 & 94.85/90.97 & 46.13/94.64 & 92.34/89.30 & 29.12/\underline{91.93} & 88.12/81.35 & 33.07/\underline{87.54} & 68.54/78.68 \\
GPT-5 & 84.86/82.10 & 38.42/93.06 & 94.25/91.07 & 43.64/94.58 & 92.14/89.29 & 26.36/91.43 & \underline{88.64}/81.92 & 31.34/87.46 & \underline{70.06}/79.02 \\
NLLB-54.5B & -- & 38.05/86.86 & 88.51/83.17 & 43.23/90.55 & 88.32/87.16 & 25.17/82.37 & 80.61/75.64 & 20.72/65.31 & 47.49/60.65 \\
TranslateGemma-4B & 75.97/73.03 & 27.71/82.77 & 85.87/82.78 & 33.42/88.43 & 86.14/86.65 & 17.71/78.13 & 77.90/73.45 & 23.37/80.34 & 64.54/75.85 \\
TranslateGemma-12B & 84.16/81.63 & 31.05/91.20 & 93.63/90.32 & 35.45/91.98 & 89.30/88.01 & 21.56/87.25 & 85.56/81.38 & 26.75/85.95 & 68.32/79.50 \\
TranslateGemma-27B & \underline{86.08}/\underline{83.31} & 32.44/93.46 & \underline{95.14}/\underline{91.67} & 38.31/93.61 & 91.21/88.83 & 23.47/89.89 & 87.60/\underline{82.66} & 28.75/87.31 & 68.76/\underline{80.11} \\
MiLMMT-46-1B-v0.1 & 76.14/73.08 & 35.07/86.35 & 89.33/85.11 & 40.61/90.68 & 90.00/88.06 & 22.12/80.82 & 81.06/75.85 & 26.94/79.80 & 64.61/75.32 \\
MiLMMT-46-4B-v0.1 & 83.27/79.93 & 39.56/92.05 & 93.91/89.79 & 43.93/93.61 & 92.25/89.26 & 26.41/88.91 & 87.45/80.91 & 30.96/85.05 & 67.87/78.57 \\
MiLMMT-46-12B-v0.1 & 85.09/81.75 & 41.24/93.46 & 94.88/90.88 & 45.45/94.37 & \underline{92.76}/\underline{89.52} & 28.27/91.00 & 88.60/82.05 & 32.44/86.65 & 68.95/79.57 \\
MiLMMT-46-1B-v1.0 & 79.01/75.82 & 33.99/88.13 & 91.11/86.87 & 39.52/91.35 & 90.90/88.34 & 21.72/82.77 & 83.21/78.09 & 26.80/81.52 & 67.08/77.01 \\
MiLMMT-46-4B-v1.0 & 85.86/82.20 & 37.28/93.29 & 95.09/91.07 & 42.68/93.94 & 92.74/89.38 & 25.33/89.83 & 88.45/82.29 & 30.56/86.60 & 69.93/79.80 \\
MiLMMT-46-12B-v1.0 & \textbf{87.88}/\textbf{84.06} & 38.04/\textbf{94.68} & \textbf{96.09}/\textbf{92.19} & 44.04/\underline{94.70} & \textbf{93.18}/\textbf{89.64} & 26.79/91.83 & \textbf{89.62}/\textbf{83.41} & 31.68/\textbf{88.18} & \textbf{71.20}/\textbf{80.79} \\
\hline
\end{tabular}%
}
\caption{Translation performance on WMT24++ and FLORES+. WMT24++ QE cells report reference-free XCOMET / COMETKiwi. For FLORES+, Ref. cells report spBLEU / reference-based XCOMET, while QE cells report reference-free XCOMET / COMETKiwi. The best and second-best scores for each metric within each language-coverage block are shown in \textbf{bold} and \underline{underlined}, respectively. Detailed reference-free and reference-based results are provided in Appendix~\ref{app:reference_free_results} and Appendix~\ref{app:reference_based_results}, respectively.}
\label{tab:main_results}
\end{table*}

%% file: appendix_reference_tables.tex
\subsection{Reference-Based Results}\label{app:reference_based_results}
Tables~\ref{tab:flores_plus_ref_baseline_en}--\ref{tab:flores_plus_ref_ours_zh} report the per-direction reference-based FLORES+ results (spBLEU / XCOMET).

\begin{table*}[!ht]
\setlength{\tabcolsep}{2pt}
\centering
\resizebox{\textwidth}{!}{
    \begin{tabular}{c | c c c | c c | c c c c c c | c c | c c c}
Direction & TranslateGemma-4B & TranslateGemma-12B & TranslateGemma-27B & GemmaX2-28-2B & GemmaX2-28-9B & Hunyuan-MT-7B & HY-MT1.5-1.8B & HY-MT1.5-7B & HY-MT2-1.8B & HY-MT2-7B & HY-MT2-30B-A3B & Seed-X-Instruct-7B & Seed-X-PPO-7B & Tower-Plus-2B & Tower-Plus-9B & Tower-Plus-72B \\
\hline
\hline
    \texttt{en} $\rightarrow$ \texttt{ar} & 31.42 / 92.06 & 33.73 / 94.89 & 33.79 / 96.20 & 37.71 / 90.33 & 41.91 / 93.56 & 32.27 / 95.46 & 27.96 / 92.30 & 32.68 / 95.57 & 34.00 / 92.05 & 38.39 / 94.93 & 38.80 / 95.37 & 41.78 / 92.87 & 42.62 / 93.38 & - / - & - / - & - / - \\
    \texttt{ar} $\rightarrow$ \texttt{en} & 34.36 / 89.18 & 36.37 / 92.60 & 40.06 / 94.09 & 45.26 / 92.01 & 48.05 / 94.18 & 32.52 / 92.22 & 26.51 / 89.55 & 33.42 / 92.72 & 34.33 / 89.79 & 42.98 / 93.51 & 44.06 / 94.45 & 47.05 / 92.38 & 43.72 / 94.36 & - / - & - / - & - / - \\
    \hline
    \texttt{en} $\rightarrow$ \texttt{az} & 6.82 / 39.69 & 18.27 / 78.24 & 21.93 / 86.27 & - / - & - / - & - / - & - / - & - / - & - / - & - / - & - / - & - / - & - / - & - / - & - / - & - / - \\
    \texttt{az} $\rightarrow$ \texttt{en} & 20.84 / 76.12 & 23.68 / 85.74 & 26.14 / 88.98 & - / - & - / - & - / - & - / - & - / - & - / - & - / - & - / - & - / - & - / - & - / - & - / - & - / - \\
    \hline
    \texttt{en} $\rightarrow$ \texttt{bn} & 24.35 / 87.37 & 24.81 / 91.23 & 26.10 / 92.14 & 31.60 / 88.23 & 33.13 / 90.80 & 24.89 / 92.34 & 22.67 / 88.91 & 25.77 / 92.72 & 25.82 / 88.82 & 28.90 / 92.54 & 27.63 / 93.28 & - / - & - / - & - / - & - / - & - / - \\
    \texttt{bn} $\rightarrow$ \texttt{en} & 28.66 / 85.75 & 31.33 / 89.90 & 34.53 / 91.73 & 39.77 / 89.28 & 42.80 / 91.36 & 25.16 / 85.21 & 21.36 / 85.78 & 28.52 / 86.98 & 27.55 / 85.04 & 35.89 / 89.31 & 37.36 / 91.97 & - / - & - / - & - / - & - / - & - / - \\
\hline
    \end{tabular}}
\caption{Reference-based English-centric evaluation results (spBLEU / XCOMET) of baseline models (Part I) on the FLORES+ benchmark.}\label{tab:flores_plus_ref_baseline_en}
\end{table*}

\begin{table*}[!ht]
\setlength{\tabcolsep}{2pt}
\centering
\resizebox{\textwidth}{!}{
    \begin{tabular}{c | c c c c c | c c c c c c}
Direction & Google Translate & NLLB-54.5B & Gemini 2.5 Pro & Gemini 3 Pro & GPT-5 & MiLMMT-46-1B-v0.1 & MiLMMT-46-1B-v1.0 & MiLMMT-46-4B-v0.1 & MiLMMT-46-4B-v1.0 & MiLMMT-46-12B-v0.1 & MiLMMT-46-12B-v1.0 \\
\hline
\hline
    \texttt{en} $\rightarrow$ \texttt{ar} & 44.90 / 95.14 & 43.03 / 90.90 & 41.55 / 95.43 & 43.60 / 95.78 & 38.68 / 94.69 & 34.99 / 87.04 & 33.66 / 88.67 & 40.31 / 93.00 & 37.48 / 93.95 & 42.61 / 94.34 & 38.48 / 95.90 \\
    \texttt{ar} $\rightarrow$ \texttt{en} & 50.99 / 94.87 & 48.26 / 92.50 & 49.55 / 95.54 & 49.26 / 95.67 & 46.74 / 95.10 & 42.43 / 90.13 & 41.25 / 90.60 & 46.94 / 93.51 & 45.54 / 93.63 & 48.92 / 94.78 & 47.23 / 94.96 \\
    \hline
    \texttt{en} $\rightarrow$ \texttt{az} & 26.51 / 82.61 & 24.63 / 79.16 & 27.41 / 90.16 & 28.07 / 90.47 & 25.85 / 89.46 & 22.27 / 76.34 & 21.73 / 79.44 & 25.64 / 86.38 & 24.96 / 88.19 & 26.95 / 88.76 & 25.89 / 90.61 \\
    \texttt{az} $\rightarrow$ \texttt{en} & 33.24 / 90.76 & 31.94 / 87.01 & 31.77 / 91.71 & 32.12 / 91.82 & 29.73 / 91.43 & 26.22 / 84.35 & 25.31 / 86.04 & 29.18 / 89.82 & 28.23 / 90.22 & 30.49 / 90.89 & 29.24 / 91.71 \\
    \hline
    \texttt{en} $\rightarrow$ \texttt{bn} & 37.55 / 88.41 & 36.04 / 85.21 & 33.47 / 90.64 & 32.86 / 92.28 & 28.46 / 89.72 & 27.62 / 83.29 & 25.69 / 86.41 & 31.66 / 89.69 & 28.59 / 92.01 & 32.73 / 91.35 & 29.01 / 93.51 \\
    \texttt{bn} $\rightarrow$ \texttt{en} & 44.97 / 91.90 & 42.22 / 89.21 & 44.35 / 92.95 & 44.18 / 93.28 & 41.50 / 93.03 & 36.85 / 86.34 & 36.20 / 87.79 & 41.63 / 90.72 & 40.39 / 91.37 & 43.04 / 91.92 & 41.73 / 92.52 \\
    \hline
    \texttt{en} $\rightarrow$ \texttt{bg} & 53.44 / 95.44 & 49.98 / 92.10 & 51.39 / 96.37 & 52.29 / 96.67 & 47.61 / 95.70 & 45.61 / 90.51 & 43.85 / 91.80 & 50.79 / 95.03 & 47.97 / 95.89 & 51.87 / 95.99 & 47.29 / 96.79 \\
\hline
    \end{tabular}}
\caption{Reference-based English-centric evaluation results (spBLEU / XCOMET) of baseline models (Part II) and MiLMMT models on the FLORES+ benchmark.}\label{tab:flores_plus_ref_ours_en}
\end{table*}

\begin{table*}[!ht]
\setlength{\tabcolsep}{2pt}
\centering
\resizebox{\textwidth}{!}{
    \begin{tabular}{c | c c c | c c | c c c c c c | c c | c c c}
Direction & TranslateGemma-4B & TranslateGemma-12B & TranslateGemma-27B & GemmaX2-28-2B & GemmaX2-28-9B & Hunyuan-MT-7B & HY-MT1.5-1.8B & HY-MT1.5-7B & HY-MT2-1.8B & HY-MT2-7B & HY-MT2-30B-A3B & Seed-X-Instruct-7B & Seed-X-PPO-7B & Tower-Plus-2B & Tower-Plus-9B & Tower-Plus-72B \\
\hline
\hline
    \texttt{zhs} $\rightarrow$ \texttt{ar} & 20.57 / 85.48 & 22.40 / 90.53 & 23.94 / 92.21 & 23.78 / 85.41 & 28.27 / 90.53 & 23.10 / 91.34 & 20.05 / 86.54 & 22.37 / 91.54 & 22.22 / 87.65 & 24.07 / 92.15 & 26.50 / 92.09 & 26.36 / 88.77 & 26.77 / 90.74 & - / - & - / - & - / - \\
    \texttt{ar} $\rightarrow$ \texttt{zhs} & 23.42 / 81.50 & 26.12 / 87.23 & 28.87 / 88.59 & 29.88 / 83.94 & 32.97 / 87.55 & 22.93 / 88.19 & 22.40 / 84.78 & 24.11 / 88.83 & 23.76 / 84.67 & 24.10 / 89.41 & 30.56 / 90.10 & 31.85 / 86.49 & 27.59 / 87.27 & - / - & - / - & - / - \\
    \hline
    \texttt{zhs} $\rightarrow$ \texttt{az} & 4.75 / 34.69 & 15.04 / 71.10 & 17.28 / 79.83 & - / - & - / - & - / - & - / - & - / - & - / - & - / - & - / - & - / - & - / - & - / - & - / - & - / - \\
    \texttt{az} $\rightarrow$ \texttt{zhs} & 18.51 / 68.43 & 21.47 / 78.90 & 23.13 / 81.09 & - / - & - / - & - / - & - / - & - / - & - / - & - / - & - / - & - / - & - / - & - / - & - / - & - / - \\
    \hline
    \texttt{zhs} $\rightarrow$ \texttt{bn} & 16.84 / 83.05 & 18.08 / 88.10 & 20.66 / 89.69 & 20.33 / 84.02 & 23.41 / 88.28 & 18.16 / 87.60 & 16.49 / 83.25 & 17.86 / 88.29 & 17.70 / 84.48 & 18.86 / 90.62 & 20.53 / 91.40 & - / - & - / - & - / - & - / - & - / - \\
    \texttt{bn} $\rightarrow$ \texttt{zhs} & 21.35 / 74.92 & 24.93 / 81.79 & 26.90 / 83.97 & 26.94 / 78.04 & 30.01 / 82.77 & 18.29 / 77.93 & 18.56 / 76.90 & 20.32 / 80.24 & 19.01 / 76.65 & 20.88 / 83.59 & 27.33 / 85.93 & - / - & - / - & - / - & - / - & - / - \\
\hline
    \end{tabular}}
\caption{Reference-based Chinese-centric evaluation results (spBLEU / XCOMET) of baseline models (Part I) on the FLORES+ benchmark.}\label{tab:flores_plus_ref_baselines_zh}
\end{table*}

\begin{table*}[!ht]
\setlength{\tabcolsep}{2pt}
\centering
\resizebox{\textwidth}{!}{
    \begin{tabular}{c | c c c c c | c c c c c c}
Direction & Google Translate & NLLB-54.5B & Gemini 2.5 Pro & Gemini 3 Pro & GPT-5 & MiLMMT-46-1B-v0.1 & MiLMMT-46-1B-v1.0 & MiLMMT-46-4B-v0.1 & MiLMMT-46-4B-v1.0 & MiLMMT-46-12B-v0.1 & MiLMMT-46-12B-v1.0 \\
\hline
\hline
    \texttt{zhs} $\rightarrow$ \texttt{ar} & 31.72 / 91.10 & 27.53 / 86.46 & 29.47 / 92.89 & 30.70 / 93.03 & 26.38 / 92.23 & 21.96 / 81.00 & 21.59 / 82.73 & 26.33 / 88.56 & 25.33 / 89.73 & 28.77 / 91.25 & 27.13 / 92.36 \\
    \texttt{ar} $\rightarrow$ \texttt{zhs} & 37.23 / 87.26 & 21.25 / 70.04 & 34.46 / 89.09 & 35.09 / 89.37 & 32.71 / 89.05 & 26.16 / 80.95 & 26.13 / 82.10 & 30.85 / 85.73 & 30.42 / 87.01 & 32.55 / 88.06 & 32.00 / 89.16 \\
    \hline
    \texttt{zhs} $\rightarrow$ \texttt{az} & 21.35 / 79.59 & 19.24 / 70.95 & 22.51 / 86.13 & 22.90 / 86.77 & 20.43 / 84.90 & 16.86 / 66.61 & 16.48 / 69.88 & 19.86 / 79.83 & 18.91 / 81.08 & 21.21 / 84.04 & 20.46 / 84.95 \\
    \texttt{az} $\rightarrow$ \texttt{zhs} & 30.41 / 79.10 & 19.32 / 64.19 & 27.77 / 82.79 & 28.12 / 82.86 & 25.42 / 83.07 & 21.06 / 70.93 & 20.83 / 73.74 & 25.17 / 79.29 & 24.55 / 80.79 & 26.56 / 81.08 & 25.76 / 82.83 \\
    \hline
    \texttt{zhs} $\rightarrow$ \texttt{bn} & 28.29 / 86.23 & 23.67 / 81.38 & 24.99 / 89.62 & 25.28 / 90.72 & 21.08 / 90.00 & 17.26 / 78.55 & 17.51 / 81.08 & 21.89 / 87.10 & 20.13 / 88.57 & 23.36 / 89.17 & 21.52 / 90.90 \\
    \texttt{bn} $\rightarrow$ \texttt{zhs} & 34.34 / 82.48 & 20.01 / 63.34 & 31.52 / 85.02 & 32.57 / 85.09 & 30.31 / 85.45 & 23.20 / 72.94 & 23.47 / 75.43 & 28.46 / 80.47 & 27.97 / 82.44 & 30.24 / 83.32 & 29.62 / 85.00 \\
    \hline
    \texttt{zhs} $\rightarrow$ \texttt{bg} & 36.71 / 94.01 & 30.60 / 87.40 & 34.66 / 95.22 & 35.53 / 95.40 & 30.27 / 95.01 & 26.84 / 85.76 & 26.25 / 87.18 & 31.48 / 92.71 & 30.43 / 93.36 & 33.28 / 94.13 & 31.90 / 94.58 \\
\hline
    \end{tabular}}
\caption{Reference-based Chinese-centric evaluation results (spBLEU / XCOMET) of baseline models (Part II) and MiLMMT models on the FLORES+ benchmark.}\label{tab:flores_plus_ref_ours_zh}
\end{table*}